\RequirePackage{fix-cm}
\documentclass[twocolumn]{svjour3}          % twocolumn
\smartqed  % flush right qed marks, e.g. at end of proof

\usepackage{cite}
\usepackage[pdftex]{graphicx}
\usepackage{ragged2e}
\usepackage[tight,footnotesize]{subfigure}
\usepackage{rotating}
\usepackage{graphicx}
\usepackage{amsmath,amssymb} % define this before the line numbering.
\usepackage{array}
\usepackage{enumitem}
\usepackage{multirow}
\usepackage{colortbl}
\usepackage{amsfonts}
\usepackage{pifont}
\usepackage{xspace}
\usepackage{etoolbox}
\usepackage{color}
\usepackage{microtype}
\usepackage{booktabs}
\usepackage{upgreek}
\usepackage[percent]{overpic}

\newcommand{\orcid}[1]{\href{https://orcid.org/#1}{\includegraphics[width=10pt]{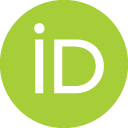}}}

\def\etal{{\em et al}}

\usepackage{hyperref}
\hypersetup{breaklinks=true,citecolor=blue, colorlinks}

\graphicspath{{./Imgs/}}
\DeclareGraphicsExtensions{.pdf,.jpg,.png}

\usepackage{silence}
\journalname{Research Article}

\begin{document}

\title{URNet: Uncertainty-aware Refinement Network for Event-based Stereo Depth Estimation}

\titlerunning{URNet}        % For running head

\author{Yifeng Cheng \orcid{0009-0005-0348-7883}        \and
  Alois Knoll \orcid{0000-0003-4840-076X} \and 
  Hu Cao \orcid{0000-0001-8225-858X}
}

\authorrunning{Yifeng Cheng \etal} % if too long for running head

\institute{
Yifeng Cheng, Alois Knoll and Hu Cao are with the Chair of Robotics, Artificial Intelligence and Real-time Systems, Technical University of Munich (TUM), Munich, Germany. 
(Email: ge87hiy@mytum.de, k@tum.de, hu.cao@tum.de). \\
Corresponding author: Hu Cao (hu.cao@tum.de).
}

\date{Received: date / Accepted: date}
% The correct dates will be entered by the editor

\maketitle

\begin{abstract}

Event cameras provide high temporal resolution, high dynamic range, and low latency, offering significant advantages over conventional frame-based cameras. In this work, we introduce an uncertainty-aware refinement network called URNet for event-based stereo depth estimation. Our approach features a local-global refinement module that effectively captures fine-grained local details and long-range global context. Additionally, we introduce a Kullback-Leibler (KL) divergence-based uncertainty modeling method to enhance prediction reliability. Extensive experiments on the DSEC dataset demonstrate that URNet consistently outperforms state-of-the-art (SOTA) methods in both qualitative and quantitative evaluations.

% Please provide 4 to 6 keywords which can be used for indexing purposes.

\end{abstract}

\section{Introduction}
\label{sec:intr}

Stereo depth estimation is a fundamental problem in computer vision, playing a vital role in applications such as autonomous driving and robotics~\cite{9233988,8237279,988771}. By triangulating corresponding points between two camera views, stereo methods can recover dense geometric information about the scene. However, conventional frame-based stereo approaches often struggle under challenging conditions such as severe motion blur and low illumination~\cite{mostafavi2021event, 9233988}.
To address these limitations, event cameras (also known as neuromorphic cameras) have gained increasing attention~\cite{6889103,4444573,9138762}. Unlike conventional RGB cameras that capture full image frames at fixed time intervals, event cameras operate asynchronously, outputting pixel-level brightness changes only when the change exceeds a certain threshold. This event-driven design enables high temporal resolution, low latency, and a wide dynamic range of approximately 120 dB, significantly reducing motion blur and over/underexposure issues~\cite{9138762,Gehrig21ral}. As a result, event-based stereo vision, which uses a pair of event cameras, has emerged as a promising approach for robust and real-time depth estimation in fast-moving and high dynamic range environments.

However, event-based stereo vision presents several inherent challenges. Sparse event streams result in limited density, particularly in low-texture and static regions, while the asynchronous event measurements often fail to capture smooth surfaces and fine local details~\cite{mostafavi2021event}. Moreover, sensor noise and motion ambiguity introduce additional uncertainty, complicating accurate depth estimation in complex environments. To address these issues, recent works have proposed diverse solutions~\cite{mostafavi2021event, nam2022stereo, tes24eccv, bartolomei2024lidar, 9878658}. For example, Se-cff~\cite{nam2022stereo} employs a concentration network to improve prediction precision. EI-Stereo~\cite{mostafavi2021event} fuses events and intensity images to mitigate event sparsity and motion-induced noise via an event-intensity recycling network. TemporalEventStereo~\cite{tes24eccv} leverages temporal event flow to enhance stereo matching consistency. Furthermore, Bartolomei et al.~\cite{bartolomei2024lidar} augmented sparse events with hallucinated measurements derived from sparse light detection and ranging (LiDAR) depth, while Zhang et al.~\cite{9878658} modeled the spatiotemporal continuity of events using continuous-time convolution. Despite these advancements, a unified framework that jointly refines depth estimation and incorporates uncertainty-aware optimization remains lacking.

\begin{figure}[t!]
    \centering
    \includegraphics[width=\linewidth, trim=10cm 9cm 12cm 9cm, clip]{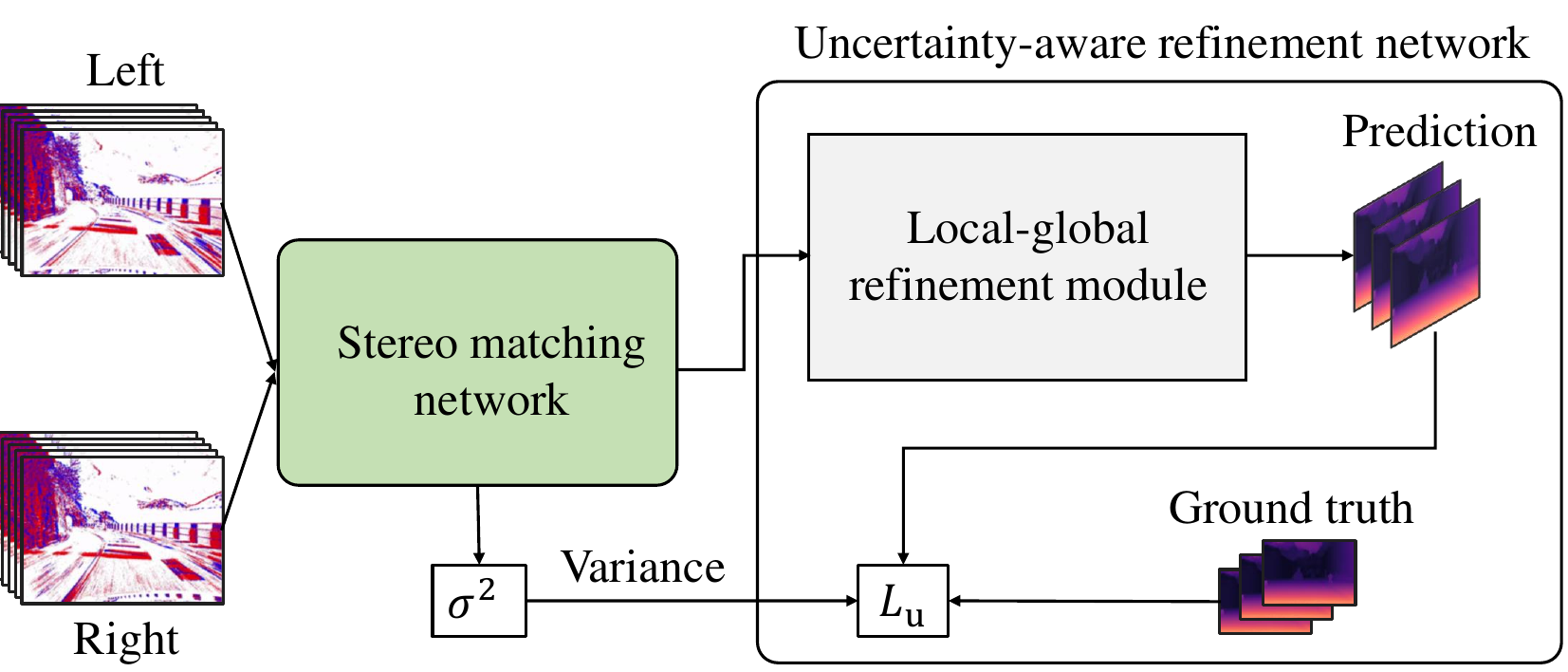}
    \caption{Overview of our uncertainty-aware refinement network (URNet). The proposed framework processes stereo event streams to estimate depth. An initial disparity map is generated by a stereo matching network, followed by refinement through a local-global refinement module that captures both fine-grained details and global context. The final predictions are supervised using an uncertainty-aware loss, leading to more robust and accurate results. Here, $\sigma^2$ represents the per-pixel prediction variance and $L_{\text{u}}$ denotes the uncertainty-aware loss.}
    \label{fig:network}
\end{figure}

In this paper, we propose URNet, a novel framework that substantially improves event-based stereo depth estimation by combining a local-global refinement module with uncertainty-aware learning. As illustrated in Fig.~\ref{fig:network}, the refinement module enables multi-scale disparity enhancement by capturing both fine-grained local details and global contextual cues. To model uncertainty, the network predicts a per-pixel variance alongside disparity, and this uncertainty modulates the loss function via  Kullback-Leibler (KL) divergence, reducing the impact of unreliable predictions. Extensive experiments on the DSEC dataset~\cite{Gehrig21ral} show that the URNet consistently outperforms state-of-the-art (SOTA) methods across multiple metrics, including mean absolute error (MAE), root-mean-square error (RMSE), and pixel error rates.
Specifically, our contributions can be summarized as follows:
\begin{enumerate}[label=\arabic*)]  % Customized enumeration style
    \item We propose an uncertainty-aware refinement network for event-based stereo depth estimation.
    
    \item A local-global refinement module combined with KL divergence-based uncertainty modeling is modeled to improve depth estimation accuracy. 
    
    \item We validate URNet on the DSEC dataset, showing consistent improvements over SOTA methods in both qualitative and quantitative evaluations. Our results highlight the effectiveness of combining refinement techniques with uncertainty modeling for enhanced event-based stereo depth estimation.
\end{enumerate}

\section{Related Work}\label{sec:related_work}

\subsection{Event Cameras in Vision Perception}
Event cameras have transformed vision perception in autonomous driving, offering low latency, high temporal resolution, and a broad dynamic range, making them particularly effective for handling rapid motion and challenging lighting conditions. Recent studies have explored their applications in this domain, such as steering angle prediction using deep learning~\cite{Maqueda2018cvpr} and neuromorphic visual odometry for vehicle positioning~\cite{8961878}. More comprehensive reviews of event-based neuromorphic vision are available in Refs.~\cite{9138762,9129849}. Additionally, advancements such as low-latency automotive vision systems~\cite{Gehrig24nature} further showcase their effectiveness. Moreover, Cao et al.~\cite{9546775} proposed a fusion-based feature attention gate component (FAGC) to enhance vehicle detection by effectively combining event data with grayscale images. Recently, Cao et al.~\cite{cao2024embracing} introduced a hierarchical feature refinement network (FRN) that used cross-modality adaptive feature refinement (CAFR) modules to improve event-frame fusion for object detection. Another cross-modal fusion strategy, such as the mixed attention mechanism presented in Ref.~\cite{zhang2023cmx} for RGB-X segmentation, also offered valuable insights for integrating event and frame data in visual perception tasks. However, the sparse nature of event data pose challenges for reliable depth estimation, requiring robust algorithms. Datasets such as DSEC~\cite{Gehrig21ral} provide essential benchmarks for real-world autonomous driving scenarios.

% For two-column wide figures use
\begin{figure*}[t!]
\centering
  \includegraphics[width=\linewidth, trim=8.6cm 10.2cm 6.5cm 5.8cm, clip]{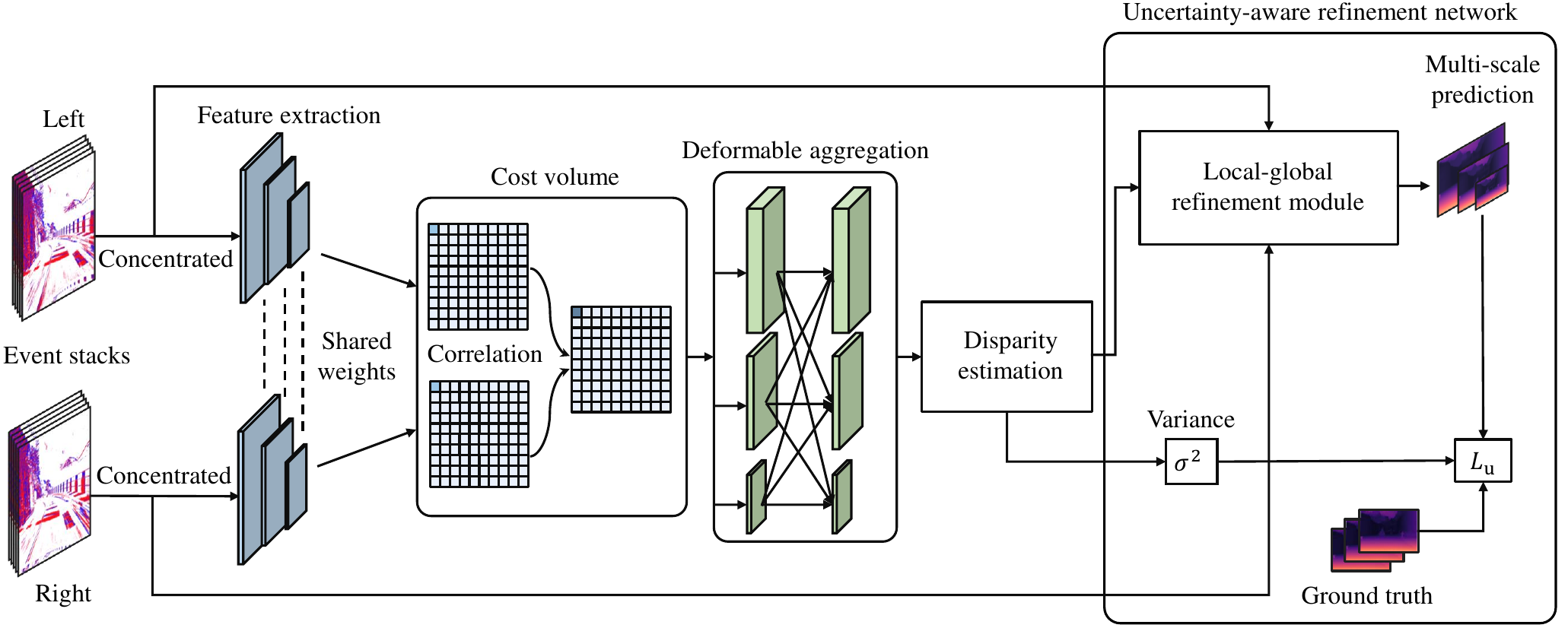}
  \caption{Overview of our URNet. The model begins by receiving event stacks from the left and right event cameras, followed by feature extraction using a shared multi-scale encoder. Cost volumes are then constructed by correlating the feature maps to capture potential disparities across multiple resolutions. To enhance structural coherence, we introduce a deformable aggregation module that refines these correlations. Finally, an uncertainty-aware refinement network is employed to further improve prediction accuracy.}
  \label{fig:model_architecture}       % Give a unique label
\end{figure*}

\subsection{Stereo Depth Estimation}
Stereo depth estimation aims to derive depth from stereo image pairs. To benchmark stereo vision methods under real-world driving conditions, the KITTI dataset~\cite{6248074} has been widely adopted as a standard benchmark. Traditional methods rely on block matching and cost volume filtering to infer scene geometry~\cite{4359315}. With the rise of deep learning, the convolutional neural network (CNN)-based architectures have replaced handcrafted algorithms~\cite{8237279, guo2019group}. One notable example is PSMNet~\cite{8578665}, which introduces spatial pyramid pooling and a 3D CNN to improve depth estimation by leveraging global contextual information. LEAStereo~\cite{cheng2020hierarchical} applies a hierarchical neural architecture search (NAS) framework to jointly optimize feature extraction and cost volume processing, achieving SOTA results with reduced computational complexity. CREStereo~\cite{li2022practical} further enhances practical stereo matching through a cascaded recurrent refinement strategy with adaptive correlation. More recently, CroCo v2~\cite{croco_v2} has improved self-supervised pre-training for stereo matching by leveraging large-scale real-world data and introducing relative positional embeddings into vision transformers. IGEV-Stereo~\cite{xu2023iterative} proposes an iterative geometry encoding volume that captures both local correlations and non-local geometric context. Recent work has begun leveraging vision foundation models (VFMs) to enhance stereo matching~\cite{10693503}. Recently, Stereo Anything~\cite{guo2024stereo} has explored building a general-purpose stereo matching model by leveraging large-scale synthetic data and domain distillation, demonstrating strong cross-domain generalization ability.
Stereo Anywhere~\cite{Bartolomei_2025_CVPR} further improves zero-shot generalization by combining monocular depth priors from foundation models with stereo geometry in a dual-branch architecture.

\subsection{Event-based Depth Estimation}

Event-based depth estimation takes advantage of the high temporal resolution and dynamic range of event data~\cite{PMID:31191287,10.1007/978-3-642-24028-7_62,6112233}. Early approaches focused on temporal alignment and handcrafted features~\cite{scheerlinck2018continuous}, while recent deep learning-based methods have integrated event streams with intensity images~\cite{nam2022stereo} or learned direct embeddings from event sequences~\cite{9008838}. Zhu et al.~\cite{zhu2018realtime} presented synchronizing events using camera velocity to enhance stereo matching accuracy. Zhou et al.~\cite{zhou2018semi} introduced a semi-dense 3D reconstruction method for stereo event cameras. Recent advancements have employed innovative architectures to further improve performance, such as LiDAR-event fusion with hallucination techniques~\cite{bartolomei2024lidar} and temporal aggregation via stereoscopic flow~\cite{tes24eccv}. Additionally, combining both events and intensity images can improve stereo depth estimation accuracy by leveraging the strengths of each modality~\cite{mostafavi2021event}.

While our work builds upon recent advances in event-based stereo depth estimation, it introduces several key innovations. In contrast to the EI-Stereo~\cite{mostafavi2021event}, which relies on event-intensity fusion, our method operates solely on event data and achieves enhanced performance by learning to predict per-pixel confidence in a probabilistic framework. Compared with the SCSNet~\cite{cho2022selection}, which utilizes modality similarity but depends on handcrafted fusion heuristics, our architecture employs fully learnable modules for feature aggregation. Se-cff~\cite{nam2022stereo} focuses on an attention-based event concentration mechanism for event representation, whereas our method introduces two core components: a local-global refinement module that captures both short-range and long-range context, and an uncertainty-aware loss based on KL divergence to improve reliability under noisy conditions. In summary, our URNet integrates refinement and uncertainty modeling in a unified framework, leading to more stable disparity predictions.

\section{Method}\label{sec:model}

In this section, we provide a detailed explanation of our proposed URNet. While inspired by the event representation and encoding design in Se-cff~\cite{nam2022stereo}, our method introduces key innovations in both refinement and uncertainty modeling. As illustrated in Fig.~\ref{fig:model_architecture}, the model begins by generating a series of multi-density event stacks. A shared-weight multi-scale encoder is then used to extract features, which are subsequently correlated to construct cost volumes. These volumes are refined through a deformable aggregation module and an uncertainty aware refinement network to produce the final outputs. 
% Additionally, we incorporate uncertainty estimation into the learning objectives to further enhance performance (Section~\ref{sec:learning_objectives}).

\subsection{Event Representation}\label{sec:data_preparation}

\begin{figure}[t!]
    \centering
    \includegraphics[width=\linewidth, trim=12.5cm 13cm 15.5cm 7.4cm, clip]{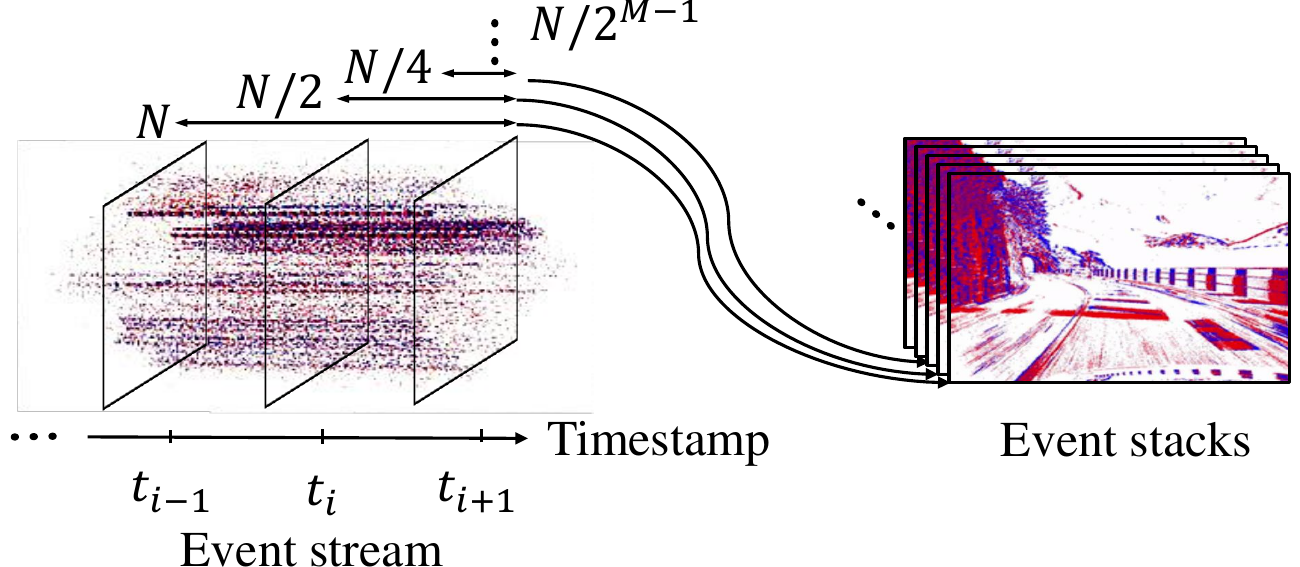}
    \caption{Illustration of the event representation. Raw event streams are transformed into structured stacks using timestamp-aligned sampling. Here, $N$ denotes the temporal length of the event sampling window, and $M$ is the number of scales used in the multi-scale event representation.}
    \label{fig:event_preparation}
\end{figure}

As illustrated in Fig.~\ref{fig:event_preparation}, each event is characterized by its pixel coordinates $(x, y)$, timestamp $t$, and polarity $p$. To ensure temporal alignment, we first discard events whose timestamps fall outside the valid range. We then apply a simple stacking-by-number strategy to collect a fixed number of events, ensuring the inclusion of both positive and negative polarities. Each pixel in the resulting stack is initialized to a default value and updated based on incoming polarity signals. For data augmentation during training, we employ random cropping and vertical flipping, while zero-padding is applied during testing to maintain consistent input dimensions. Additionally, timestamps can be sampled at user-defined intervals to skip redundant frames and enhance computational efficiency. By converting asynchronous event streams into structured representations, we preserve essential spatio-temporal information critical for accurate depth estimation. The resulting event stacks are subsequently processed by a concentration network~\cite{nam2022stereo} to produce a representation with unified channel dimensionality for downstream feature extraction and estimation tasks.

\begin{figure*}[t!]
\centering
  \includegraphics[width=\linewidth, trim=13.5cm 9.8cm 9cm 10cm, clip]{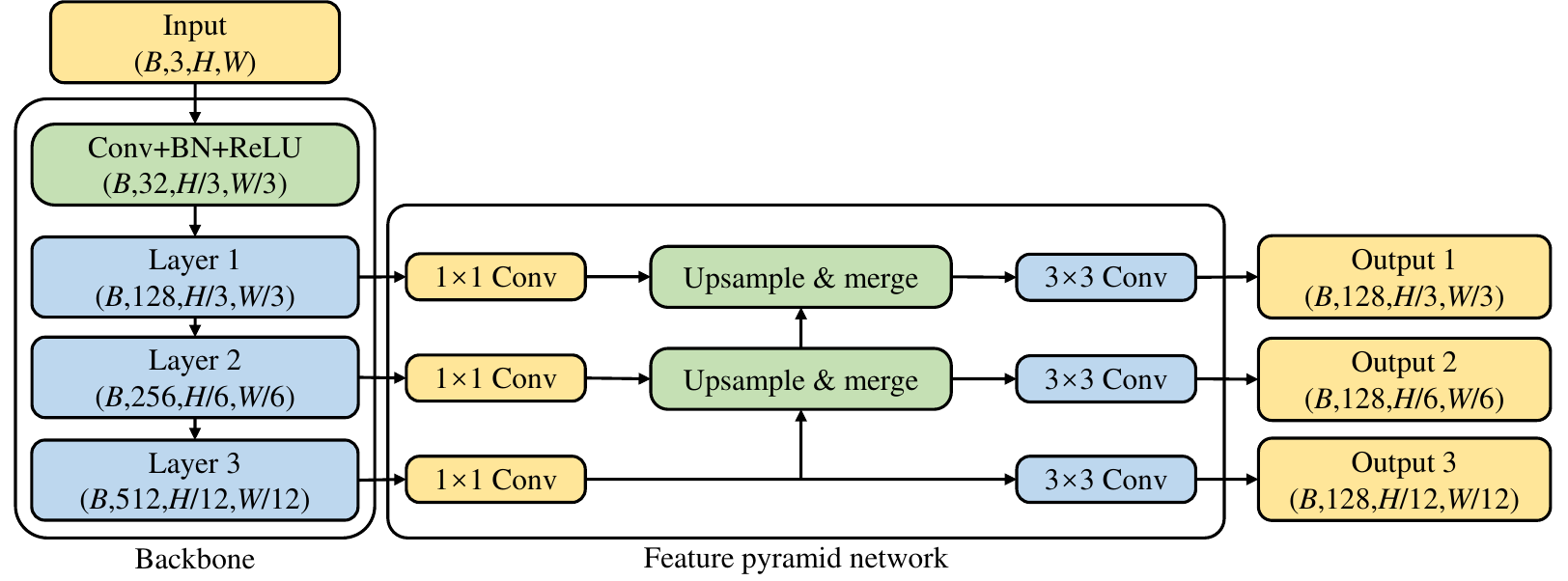}
  \caption{Architecture of the feature extraction. It consists of a backbone combined with a feature pyramid network (FPN), enabling the extraction of multi-scale features from the input event stacks. Conv, BN, and ReLU denote convolution, batch normalization, and rectified linear unit. $B$, $H$, and $W$ denote the batch size, image height, and image width.}
  \label{fig:feature_extractor}       % Give a unique label
\end{figure*}

\subsection{Feature Extraction}\label{sec:feature_extraction}

As shown in Fig.~\ref{fig:feature_extractor}, we adopt ResNet~\cite{7780459} for feature extraction because of its simplicity and proven effectiveness across a wide range of computer vision tasks. The feature extractor is shared between the left and right event cameras to ensure consistent representation learning and reduce model complexity. The ResNet-based architecture comprises three main layers, each progressively reducing the spatial resolution while capturing increasingly deeper semantic features. To better handle the sparse and irregular nature of event data, we incorporate deformable convolutions~\cite{dai17dcn} in the final layer, allowing the network to adaptively adjust its receptive fields for more robust feature extraction.
Moreover, we integrate a feature pyramid network (FPN)~\cite{8099589} to recover fine-grained details across multiple scales. FPN employs lateral connections and a top-down pathway to aggregate features from different layers, resulting in rich and coherent multi-scale representations. By merging high-level semantic features with low-level spatial information, the FPN effectively complements ResNet’s downsampling process, enhancing the robustness of the feature extraction stage for handling varying object sizes and resolutions.

\begin{figure}[t!]
  \centering
  \includegraphics[width=\linewidth, trim=12.5cm 8.5cm 15cm 6.5cm, clip]{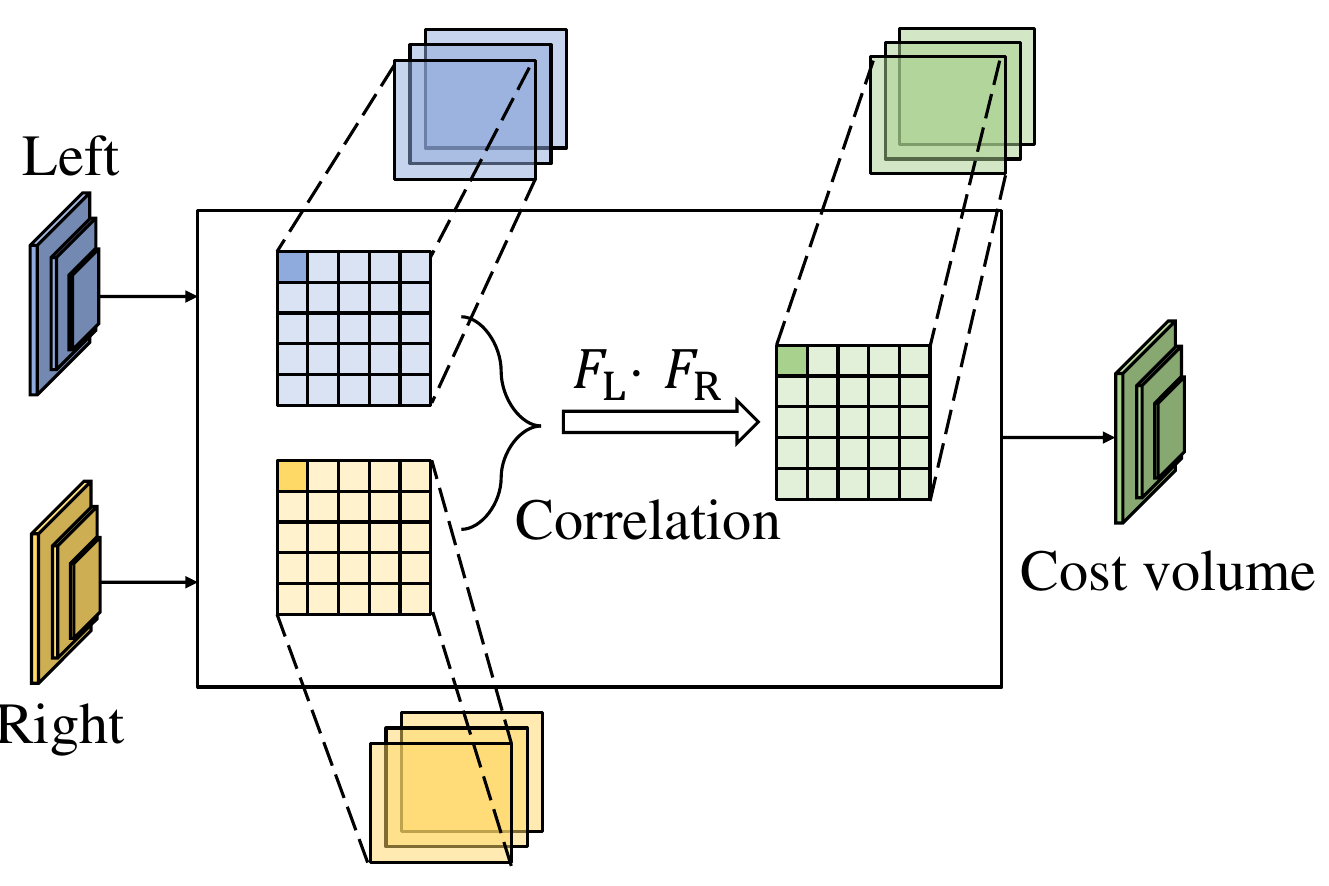}
  \caption{Illustration of the cost volume. The left feature map \(F_{\text{L}}\) and right feature map \(F_{\text{R}}\) are correlated using dot product operations across multiple pyramid levels to construct multi-scale cost volumes. }
  \label{fig:cost_volume}      
\end{figure}

\subsection{Cost Volume}\label{sec:cost_volume_construction}

As illustrated in Fig.~\ref{fig:cost_volume}, the cost volume is constructed by efficiently computing feature correspondences using a dot product correlation method~\cite{DFIB15}, which performs element-wise multiplication between the left and right feature maps. Compared to subtraction-based and concatenation-based methods~\cite{guo2019group}, dot product correlation offers better geometric consistency and lower computational cost, making it particularly well-suited for real-time stereo matching applications.
Specifically, the correlation between the left feature map \(F_{\text{L}}\) and the right feature map \(F_{\text{R}}\) is defined as
\begin{equation}
\text{Correlation}(p, d) = \frac{1}{C}\sum_{c=1}^{C} F_{\text{L}}(p,c) \cdot F_{\text{R}}(p + d, c),
\end{equation}
where \(p\) denotes the pixel location, \(d\) represents the disparity level, and \(C\) is the number of feature channels. \(F_{\text{L}}(p,c)\) and \(F_{\text{R}}(p + d,c)\) are the feature values at channel \(c\) for the left and right feature maps, respectively. Concretely, the left feature map is shifted horizontally by \(d\) pixels before performing the dot product with the right feature map. This operation produces a 4D cost volume tensor.
To effectively handle multi-scale features, we adopt a cost volume pyramid strategy. At each level of the pyramid, the disparity range is proportionally scaled to align with the corresponding spatial resolution.

\begin{figure}[t!]
  \centering
  \includegraphics[width=0.95\linewidth, trim=13.3cm 6.5cm 17.5cm 6cm, clip]{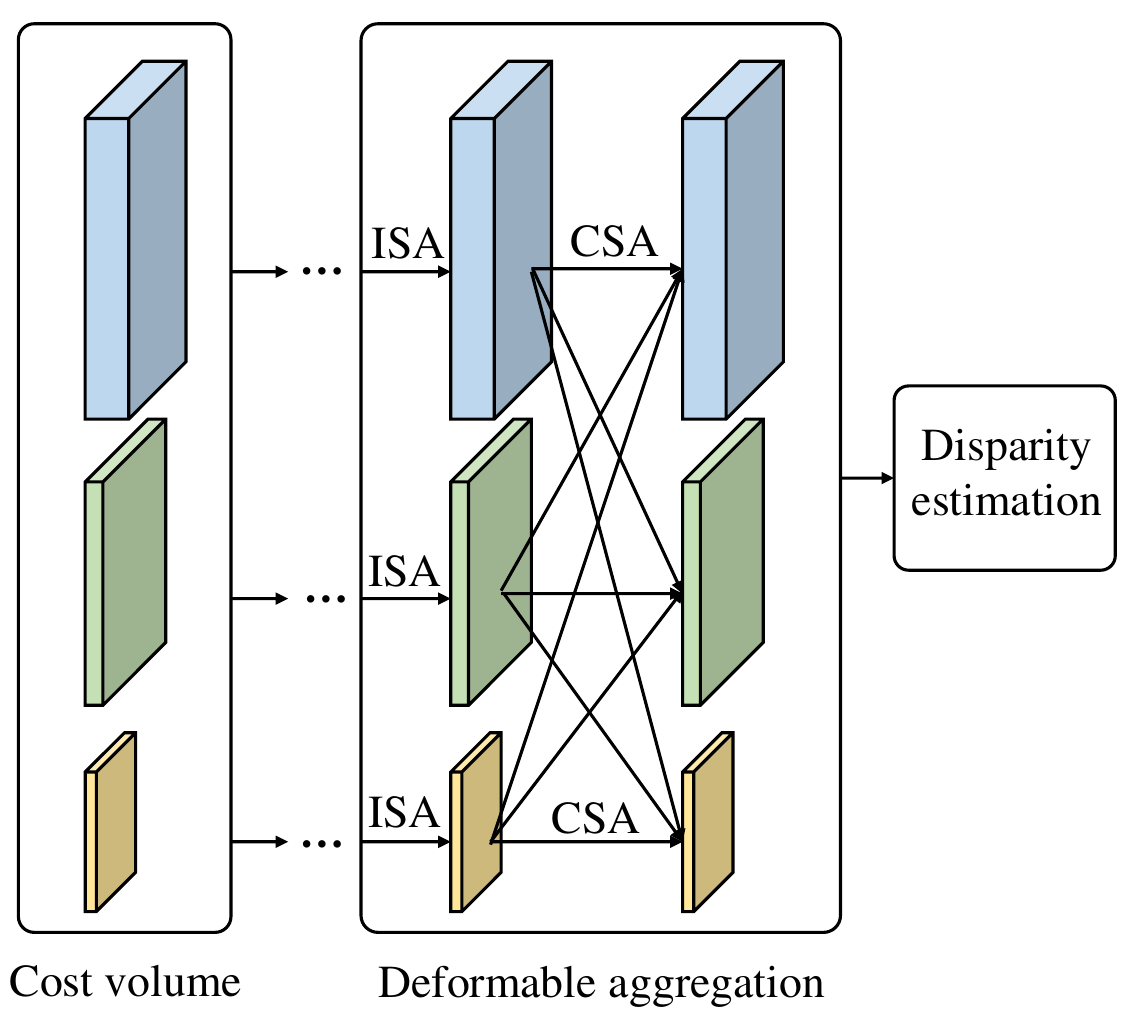}
  \caption{Illustration of the deformable aggregation. The cost volume is refined through a combination of intra-scale aggregation (ISA) and cross-scale aggregation (CSA).}
  \label{fig:aggregation}      
\end{figure}

\subsection{Deformable Aggregation}\label{sec:aggregation}

While the cost volume encodes disparity-dependent feature similarities, its raw representation often remains noisy and structurally inconsistent, especially when dealing with sparser event data. To address this, we introduce a deformable aggregation module based on deformable convolutions, which dynamically adjusts the receptive fields to better capture non-uniform spatial patterns~\cite{xu2020aanet,dai17dcn}. This flexibility enables the network to focus on relevant regions, leading to more accurate event-based stereo matching.
As demonstrated in Fig.~\ref{fig:aggregation}, our aggregation module incorporates two complementary operations: intra-scale aggregation (ISA) and cross-scale aggregation (CSA). ISA operates within each resolution level, where deformable convolutions are employed to enhance local consistency. By dynamically adjusting receptive fields, ISA effectively mitigates the challenges posed by the sparse event data. In contrast, CSA promotes multi-scale interaction by transforming and integrating features across different resolutions through a combination of upsampling, downsampling, and channel alignment. To further improve adaptability, the aggregation process is structured in multiple stages, progressively introducing additional deformable convolutions in the later stages. Furthermore, intermediate supervision is applied at various scales to enforce feature consistency.

\begin{figure*}[t]
  \centering
  \includegraphics[width=\linewidth, trim=9.9cm 6cm 8.4cm 7.5cm, clip]{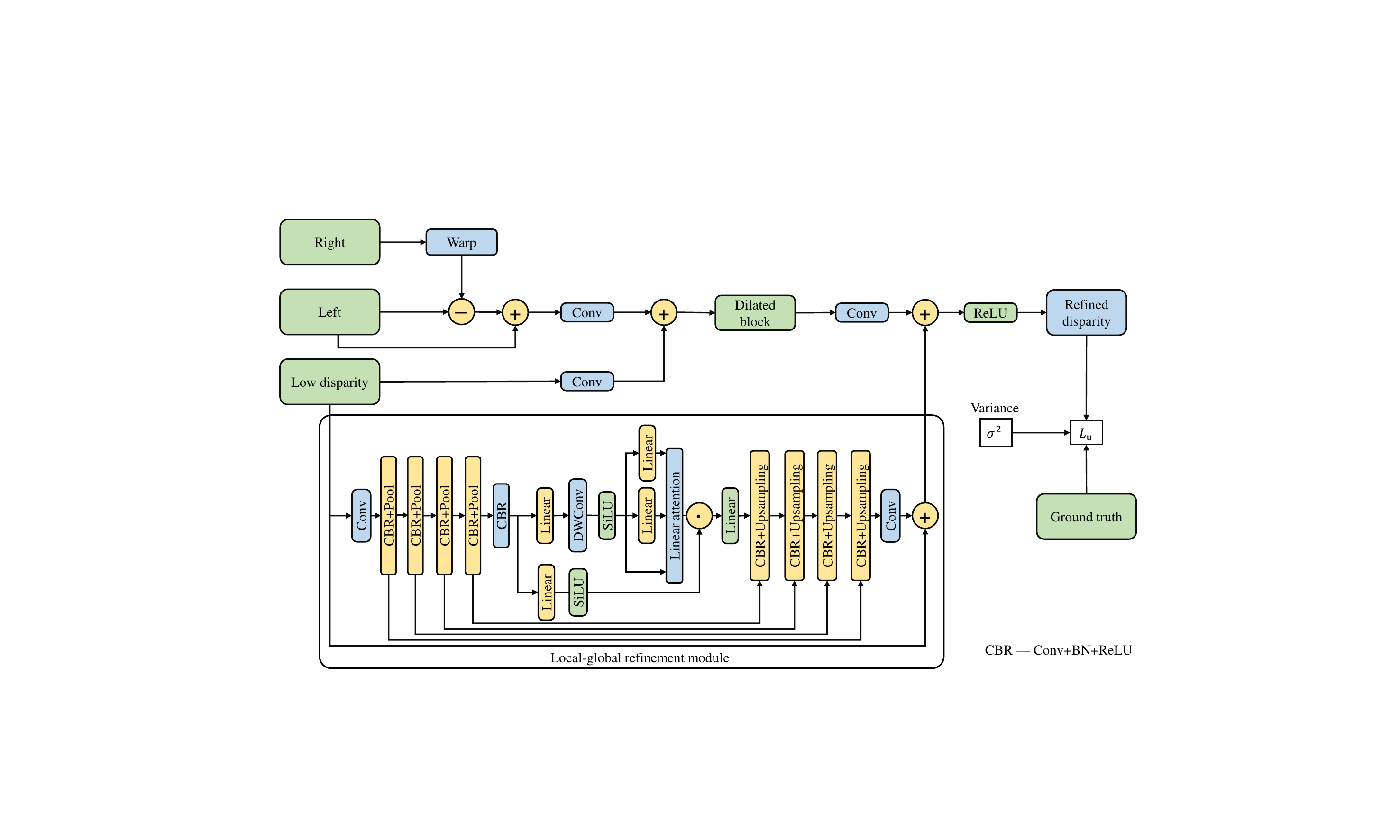}
  \caption{Structure of the proposed uncertainty-aware refinement network. It employs depthwise convolution (DWConv) and the sigmoid linear unit (SiLU) activation, with the operators $+$, $-$, and $\cdot$ denoting element-wise addition, subtraction, and multiplication.}
  \label{fig:refinementfinal}      
\end{figure*}

\subsection{Disparity Estimation}\label{sec:disparity_estimation}

Building on the aggregated cost volume, we first generate low-resolution disparity maps and progressively refine them to higher resolutions~\cite{8237279}. The initial disparity map is obtained by applying a softmax operation, effectively transforming matching costs into a probability distribution:
\begin{equation}
P(d|h,w) = \frac{\exp\bigl(C(h, w, d)\bigr)}{\sum_{d'=0}^{D-1} \exp\bigl(C(h, w, d')\bigr)},
\end{equation}
where \(P(d|h,w)\) denotes the probability of disparity \(d\) at pixel \((h,w)\), $D$ denotes the maximum disparity range, and \(C(h, w, d)\) represents the cost value at disparity \(d\). Moreover, \(d'\) is the summation index over all possible disparity candidates. Each pixel’s disparity \(\hat{d}\) is then computed as the weighted sum of all disparity candidates:
\begin{equation}
\hat{d}_{h,w} = \sum_{d=0}^{D-1} P(d|h,w)\,\cdot\,d.
\end{equation}

% To enhance accuracy, especially near object boundaries, we refine the disparity maps across multiple resolutions. This coarse-to-fine strategy efficiently processes the cost volume at different scales, yielding a high-resolution disparity map that accurately captures scene geometry and enables robust, high-quality depth estimation.

% \begin{figure}[t!]
%   \centering
%   \includegraphics[width=0.9\linewidth, trim=20cm 1cm 4cm 8cm, clip]{Imgs/linearblock.pdf}
%   \caption{Macro design of LAB, featuring an efficient attention mechanism that captures both global context and local dependencies.}
%   \label{fig:linear_block}      
% \end{figure}

\subsection{Uncertainty-aware Refinement Network}
\label{sec:disparity_refinement}

% Instead of relying on a simple convolutional refinement module~\cite{nam2022stereo}, we propose an uncertainty-aware refinement network that effectively captures both multi-scale local details and global contextual information for disparity estimation. 

\paragraph{Refinement network.}
As illustrated in Fig.~\ref{fig:refinementfinal}, the right-view input is first warped toward the left view. Then the results are processed by the initial low-resolution disparity. To further refine this initial estimation, we incorporate both a local-global refinement module and a dilated block. The local-global refinement module adopts an encoder-decoder architecture with skip connections. The encoder progressively downsamples the input using convolutional layers with batch normalization and ReLU activation to extract hierarchical features, while max pooling further reduces spatial resolution and expands the receptive field. To capture the global context, we integrate a linear attention block (LAB)~\cite{han2024demystify}, which enables efficient modeling of long-range global dependencies. The decoder then restores spatial resolution via upsampling and fuses high-resolution details from the encoder through skip connections. Finally, residual output operations are used to predict disparity correction maps, refining the initial disparity estimation by compensating for potential errors.

\paragraph{Uncertainty learning.}\label{sec:learning_objectives}

Several approaches have been explored for modeling uncertainty in deep learning-based regression tasks. Dropout-based methods~\cite{gal2016dropout} estimate uncertainty by performing stochastic forward passes during inference, but incur high computational cost. Deep ensembles~\cite{lakshminarayanan2017simple} improve prediction confidence by aggregating outputs from multiple models, at the expense of increased memory and runtime.

In this work, we introduce a more efficient uncertainty-aware learning strategy based on heteroscedastic regression to mitigate unreliable predictions in event-based stereo depth estimation. Specifically, we model each pixel’s disparity as a Gaussian distribution characterized by a learnable mean and variance. To explicitly account for prediction uncertainty, we incorporate a KL divergence loss~\cite{klloss}, which regularizes the predicted distribution against a target distribution. This formulation enables the network to express confidence in its predictions, while remaining lightweight and well-suited to dense stereo depth estimation tasks.

Specifically, the network predicts a Gaussian distribution \( P \sim \mathrm{N}(\hat{d}, \sigma^2) \) for each pixel’s disparity, where \(\hat{d}\) is the predicted mean and \(\sigma^2\) is the variance representing uncertainty. The ground truth is treated as a target Gaussian distribution \( Q \sim\mathrm{N}(d_{\text{g}}, \sigma_{\text{g}}^2)\), where \(\hat{d_{\text{g}}}\) and \(\sigma_{\text{g}}^2\) are the corresponding mean and variance of ground truth.  We measure these two distributions by minimizing the KL divergence:
\begin{equation}
D_{\text{KL}}(P \,\|\, Q) = \int_{-\infty}^{\infty} P(x) \log \frac{P(x)}{Q(x)} \, {\text{d}}x,
\end{equation}
where
\begin{align}
P(x) &= \tfrac{1}{\sqrt{2\uppi\,\sigma^2}}
\exp\!\Bigl(-\tfrac{(x - \hat{d})^2}{2\,\sigma^2}\Bigr), \\
Q(x) &= \tfrac{1}{\sqrt{2\uppi\,\sigma_{\text{g}}^2}}
\exp\!\Bigl(-\tfrac{(x - d_{\text{g}})^2}{2\,\sigma_{\text{g}}^2}\Bigr).
\end{align}
After integration and simplification, the closed-form expression for the KL divergence between two Gaussian distributions becomes

\begin{multline}
D_{\text{KL}}(P \,\|\, Q) 
= \frac{1}{2}\,\Bigl[
\tfrac{(d_{\text{g}} - \hat{d})^2}{\sigma^2} 
+ \tfrac{\sigma_{\text{g}}^2}{\sigma^2} - 1 
+ \log\tfrac{\sigma^2}{\sigma_{\text{g}}^2}
\Bigr].
\end{multline}
In practice, \(\sigma_{\text{g}}^2\) is often assumed to be zero, reflecting deterministic ground truth. Under this assumption, the KL divergence simplifies as

\begin{equation}
L_{\text{u}} 
= \tfrac{(d_{\text{g}} - \hat{d})^2}{2\,\sigma^2} 
+ \tfrac{\log\sigma^2}{2}
+ C,
\end{equation}
where \(C\) is a constant. To further stabilize training for stereo depth estimation, we let the model predict \(\log\sigma^2\) instead of \(\sigma^2\) directly. We combine the resulting uncertainty-aware KL divergence with a smooth $L_{1}$ loss~\cite{girshick2015fast} to balance accurate disparity regression with uncertainty calibration. The combined loss is formulated as:
\begin{equation}
L_{\text{u}} 
= \frac{\text{Smooth}\,L_{1}\bigl(\hat{d}, d_{\text{g}}\bigr)}{\sigma^2} 
+ \alpha \,\log\sigma^2,
\label{eq:kl_loss}
\end{equation}
where \(\alpha\) controls the penalty on large variances. We then apply this uncertainty-aware loss at multiple scales:
\begin{equation}
L_{\text{total}} = \sum_{i=1}^{N} w^i \,L_{\text{u}}^i,
\end{equation}
where \(w^i\) denotes the weighting factor for the loss at multi-scale.

% ensuring consistency across different resolutions. By adapting the predicted variance to scene complexity, this formulation balances accuracy and robustness, as our ablation studies in Section~\ref{ablation-studies} demonstrate superior performance in challenging scenarios compared to SOTA methods.

\section{Experiments}\label{sec:experiments}

\begin{figure}[t]
\centering
  \includegraphics[width=0.95\linewidth]{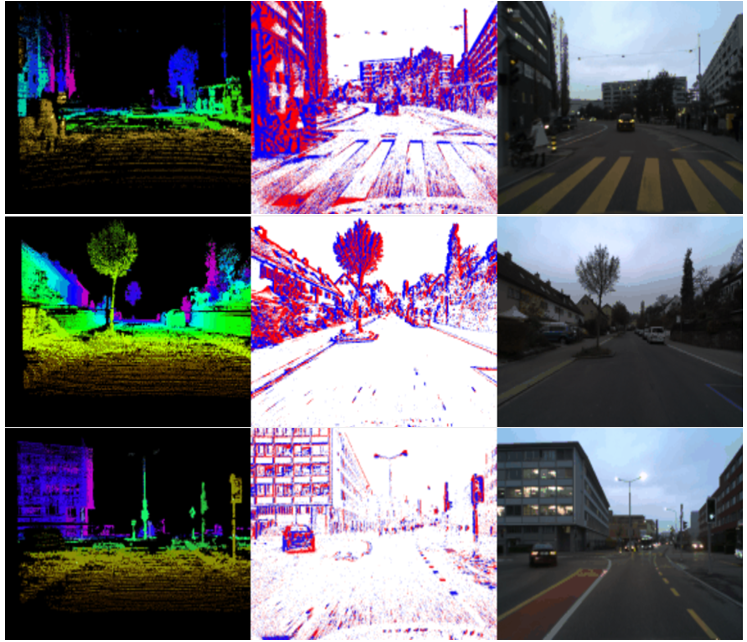}  \caption{Example samples from the DSEC dataset, showcasing light detection and ranging (LiDAR)-based disparity maps (left), corresponding event data (middle), and RGB images (right). This figure is cited from DSEC official website~\cite{Gehrig21ral}.}
  \label{fig:dsec}      
\end{figure}

\subsection{Datasets}

We evaluate our method on the DSEC dataset~\cite{Gehrig21ral}, a large-scale benchmark designed for event-based stereo vision in autonomous driving. The dataset is captured via two monochrome event cameras, two global shutter RGB cameras, and a LiDAR mounted on a vehicle. It provides asynchronous event streams, high-resolution RGB images, and LiDAR-based disparity ground truth, as illustrated in Fig.~\ref{fig:dsec}. Covering a wide range of illumination conditions, DSEC is well-suited for tasks such as depth estimation, optical flow, object detection, and segmentation. In this work, we focus on leveraging stereo event streams for depth estimation, using LiDAR-derived disparity maps as ground truth. For further details on data preprocessing and benchmark protocols, please refer to the official DSEC paper Ref.~\cite{Gehrig21ral}.

\subsection{Evaluation Metrics}

To assess the quality of the predicted disparity maps, we adopt several widely used evaluation metrics, consistent with the DSEC disparity benchmark~\cite{Gehrig21ral}. These metrics provide a comprehensive evaluation of prediction accuracy.

\paragraph{Mean absolute error (MAE).} The MAE measures the average magnitude of the absolute errors between the predicted disparity values and the ground truth. It is defined as
\begin{equation}
\text{MAE} = \frac{1}{N} \sum_{i=1}^N |p_i - g_i|,
\end{equation}
where $N$ is the total number of valid pixels, $p_i$ is the predicted disparity value at pixel $i$, and $g_i$ is the corresponding ground truth disparity value.

\paragraph{Root-mean-square error (RMSE).} The RMSE measures the standard deviation of the prediction errors, placing greater emphasis on larger errors because of the squared difference term. It is calculated as
\begin{equation}
\text{RMSE} = \sqrt{\frac{1}{N} \sum_{i=1}^N (p_i - g_i)^2}.
\end{equation}

\paragraph{1-pixel error (1PE).} The 1PE measures the proportion of pixels for which the absolute disparity error exceeds 1 pixel, indicating the frequency of noticeable prediction inaccuracies. It is formulated as
\begin{equation}
\text{1PE} = \frac{1}{N} \sum_{i=1}^N \mathbf{1}(|p_i - g_i| > 1),
\end{equation}
where $\mathbf{1}(\cdot)$ is the indicator function, which returns 1 if the condition is true and 0 otherwise.

\paragraph{2-pixel error (2PE).} Similarly, the 2PE quantifies the proportion of ground truth pixels with an absolute disparity error greater than 2 pixels:
\begin{equation}
\text{2PE} = \frac{1}{N} \sum_{i=1}^N \mathbf{1}(|p_i - g_i| > 2).
\end{equation}

% These metrics are computed using only the valid pixels as indicated by the provided mask, which ensures that occluded or undefined regions do not affect the evaluation. By combining these complementary metrics, we are able to assess not only the overall accuracy of the predicted disparity maps but also their local precision and robustness. 

\subsection{Implementation Details}

We implement our model in PyTorch and ensure reproducibility by initializing all components with fixed random seeds. The model is trained end-to-end from scratch on two NVIDIA RTX 3090 GPUs with a batch size of 8. We set the maximum disparity to 192. Optimization is performed using the Adam optimizer~\cite{Kingma2014AdamAM}, with $\beta$ parameters set to (0.900, 0.999) and a weight decay of $1.0 \times 10^{-4}$. The initial learning rate is $5.0 \times 10^{-4}$ and is gradually reduced using a cosine schedule.

\subsection{Experimental Results}\label{results-analysis}

We analyze the results of our experiments from both quantitative and qualitative perspectives. To ensure a fair comparison, we implement both our model and all the compared methods under identical settings and trained them on two NVIDIA RTX 3090 GPUs.

\begin{table*}[t!]
  \centering
  \caption{Evaluation results on the DSEC dataset~\cite{Gehrig21ral}. We report the evaluation results of the SCSNet, Se-cff, and our model, all trained under identical experimental settings and evaluated using the official DSEC evaluation platform. The DSEC baseline results are taken directly from the official benchmark~\cite{Gehrig21ral}, while the results for Se-cff with BTH and Se-cff with VSH are sourced from Ref.~\cite{bartolomei2024lidar}. (\(\downarrow\)) indicate that lower values represent better performance. LiDAR: lightlaser detection and ranging; E: event; I: image; MAE: mean absolute error; 1PE: 1-Pixel Error; 2PE: 2-Pixel Error; RMSE: root-mean-square error.}
  \renewcommand{\arraystretch}{1.3}
  \setlength{\tabcolsep}{6pt}
  \begin{tabular}{l|c|c|ccccc} 
    \toprule[1.2pt]
    Model & Venue & Modality & MAE (\(\downarrow\)) & 1PE (\%) (\(\downarrow\)) & 2PE (\%) (\(\downarrow\)) & RMSE (\(\downarrow\))  \\ 
    \midrule
    SCSNet~\cite{cho2022selection} & ECCV'22 & E + I & 0.751 & 15.616 & 4.777 & 1.891 \\
    Se-cff with BTH~\cite{bartolomei2024lidar} & ECCV'24 & E + LiDAR & 0.660 & 14.660 & 3.770 & - \\
    Se-cff with VSH~\cite{bartolomei2024lidar} & ECCV'24 & E + LiDAR & 0.650 & 13.700 & 3.600 & - \\
    DSEC Baseline~\cite{Gehrig21ral} & RAL'21 & E & 0.576 & 10.915 & 2.905 & 1.381 \\
    Se-cff~\cite{nam2022stereo} & CVPR'22 & E & 0.533 & 10.561 & 2.683 & 1.243 \\ 
    \midrule
    URNet (ours) & - & E & \textbf{0.529} & \textbf{10.054} & \textbf{2.678} & \textbf{1.234} \\ 
    \bottomrule[1.2pt]
  \end{tabular}
  \label{tab:online_results}       % Give a unique label
\end{table*}

% For two-column wide figures use
% \begin{figure*}[t!]
% \centering
%   \includegraphics[width=\linewidth]{qualitative (3).png}
%   \caption{Qualitative comparison of predictions between ours and SOTA methods.}
%   \label{fig:qualitative}       % Give a unique label
% \end{figure*}

\begin{figure*}[htbp]
\centering
\begin{overpic}[width=\linewidth]{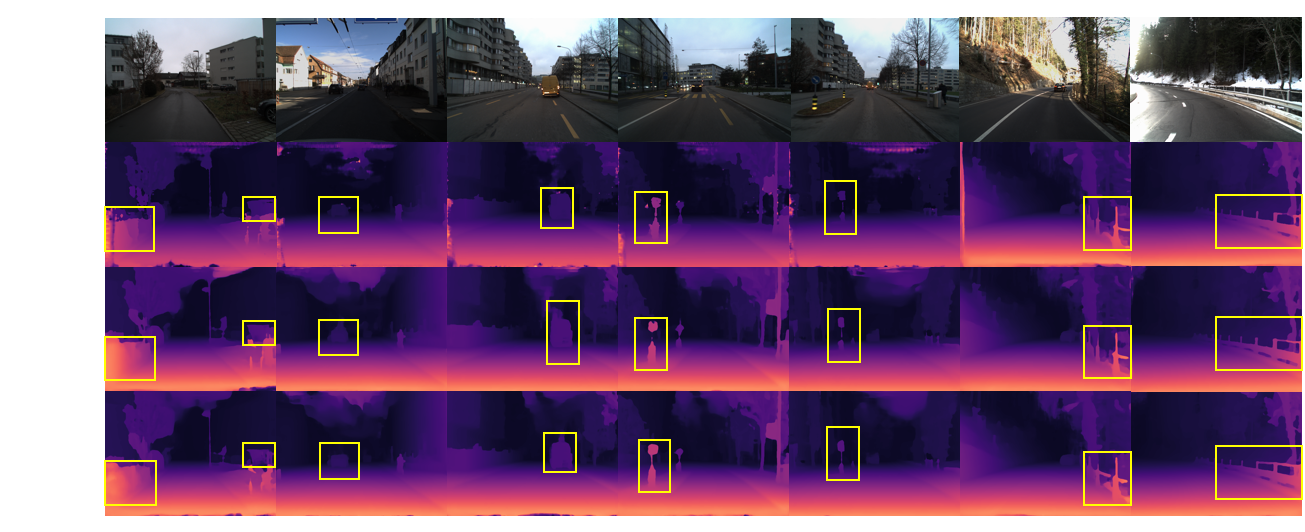}
  % 调整 (x,y) 坐标来放文字，百分比相对于整张图
  \put(-1.7,32){\small RGB image}
  \put(-1.5,23){\small SCSNet~\cite{cho2022selection}}
  \put(0,14){\small Se-cff~\cite{nam2022stereo}}
  \put(1.5,5){\small Ours}
\end{overpic}
\caption{Qualitative comparison of predictions between ours and SOTA methods.}
\label{fig:qualitative}
\end{figure*}
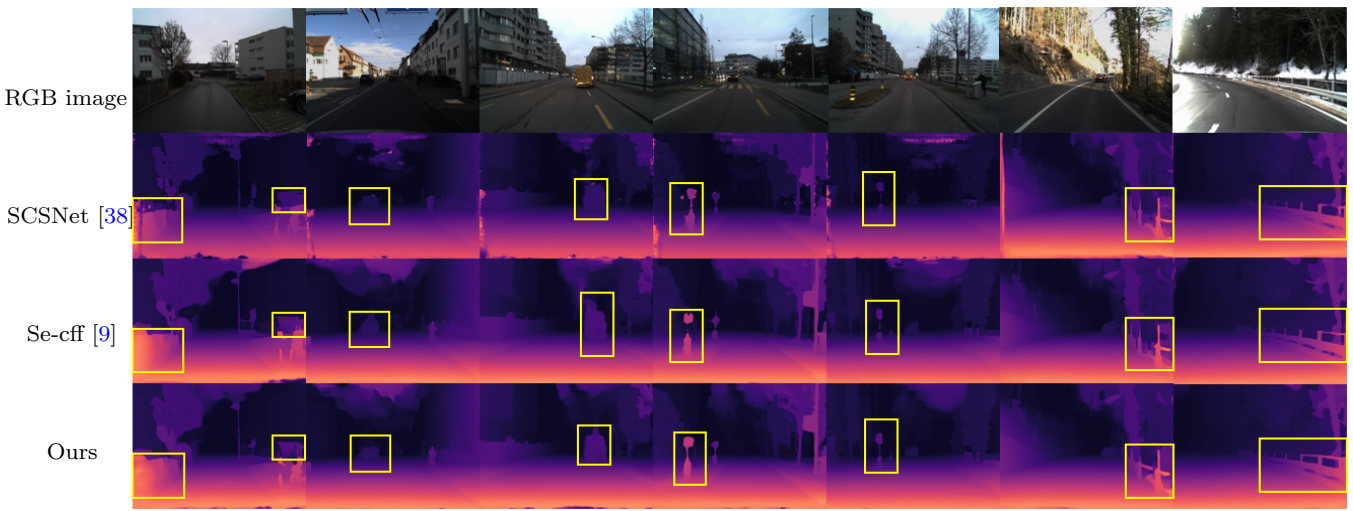

\begin{table*}[htbp]
  \centering
  \caption{Comparison of computational efficiency. All models are evaluated using official implementations on an NVIDIA RTX 3090 GPU. URNet achieves a favorable trade-off between speed, memory usage, and accuracy. (\(\uparrow\)) indicates higher is better. FLOPs: floating-point operations; Params: number of learnable parameters; Inference time: average forward pass time; FPS: frames per second; Max Mem: peak memory usage. }
  \renewcommand{\arraystretch}{1.3}
  \setlength{\tabcolsep}{6pt}
  \begin{tabular}{l|cccccc}
    \toprule[1.2pt]
    \textbf{Model} & FLOPs (G) (\(\downarrow\)) & Params (M) (\(\downarrow\)) & Inference time (s) (\(\downarrow\)) & FPS (\(\uparrow\)) & Max Mem (MB) (\(\downarrow\)) \\
    \midrule
    SCSNet & 2171.68 & 17.45 & 0.1739 & 5.75 & 1227.90 \\
    Se-cff & 446.25 & 6.92 & 0.0569 & 17.57 & 366.73 \\
    URNet (ours) & 540.49 & 7.96 & 0.0700 & 15.07 & 719.78 \\
    \bottomrule[1.2pt]
  \end{tabular}
  \label{tab:efficiency}
\end{table*}

\subsubsection{Quantitative Analysis}

We evaluate our model on the DSEC benchmark~\cite{Gehrig21ral}, comparing it with the SOTA methods. Rather than relying on the results reported in the corresponding papers, we reproduce the performance of Se-cff~\cite{nam2022stereo} and SCSNet~\cite{cho2022selection} using their official open-source implementations. It should also be noted that the official Se-cff codebase~\cite{nam2022stereo} does not include some key modules reported in the original paper, such as intensity-image processing and future event distillation. Therefore, our reproduced results of Se-cff are based on an official incomplete Se-cff codebase. All models are trained under identical experimental settings and evaluated on the same hardware environment—two NVIDIA RTX 3090 GPUs—to ensure fairness. The DSEC baseline results are obtained from the official benchmark platform~\cite{Gehrig21ral}, while the results for the variants Se-cff with BTH and Se-cff with VSH are taken directly from Ref.~\cite{bartolomei2024lidar}.
As summarized in Tab.~\ref{tab:online_results}, our proposed URNet achieves a MAE of 0.529, outperforming the second-best method Se-cff (0.533 MAE), and consistently improves across other metrics. Specifically, the 1-pixel error decreases from 10.561\% to 10.054\%, achieving a 0.507\% gain, while the RMSE improves from 1.243 to 1.234. Although the 2-pixel error reduction is marginal—from 2.683\% to 2.678\%—our method still achieves the best overall performance among all methods. Compared with the DSEC baseline (0.576 MAE), URNet reduces the MAE by 0.047. These improvements across all metrics reflect enhanced robustness and accuracy.
Note that several methods, such as Refs.~\cite{mostafavi2021event,tes24eccv}, do not provide publicly available codes and are therefore excluded from our quantitative comparison. Overall, our results demonstrate the effectiveness of URNet on real-world stereo event data.

\begin{table*}[htbp]
  \centering
  \caption{Performance comparison on the effects of local refinement module, uncertainty-aware loss using fixed \(\alpha=2.0\), and global refinement module. Here, $\alpha$ is the coefficient of the uncertainty regularization term.}
  \renewcommand{\arraystretch}{1.3}
  \setlength{\tabcolsep}{6pt}
  \begin{tabular}{l|cccc} 
    \toprule[1.2pt]
    Model configuration  & MAE (\(\downarrow\)) & 1PE (\%) (\(\downarrow\)) & 2PE (\%) (\(\downarrow\)) & RMSE (\(\downarrow\)) \\ 
    \midrule
    Baseline & 0.475 & 8.779 & 2.078 & 1.070 \\
    + Local refinement module & 0.472 & 8.686 & 2.069 & 1.066 \\
    + Local refinement module + Uncertainty & 0.454 & 7.987 & \textbf{1.995} & 1.058 \\
    + Global refinement module + Uncertainty  & 0.468 & 8.053 & 2.008 & 1.076 \\
    + Local-global refinement module + Uncertainty & \textbf{0.451} & \textbf{7.896} & 1.996 & \textbf{1.057} \\ 
    \bottomrule[1.2pt]
  \end{tabular}
  \label{tab:ablation_modules}
\end{table*}

\begin{table}[htbp]
  \centering
  \caption{Effect of varying the parameter \(\alpha\) in the loss function.}
  \renewcommand{\arraystretch}{1.1}
  \setlength{\tabcolsep}{2mm}
  \begin{tabular}{c|cccc}
    \toprule[1.2pt]
    \(\alpha\) & MAE (\(\downarrow\)) & 1PE (\%) (\(\downarrow\)) & 2PE (\%) (\(\downarrow\)) & RMSE (\(\downarrow\)) \\ 
    \midrule
    1.0 & 0.469 & 8.503 & 2.123 & 1.088 \\
    1.5 & 0.465 & 8.372 & 2.086 & 1.082 \\
    2.0 & \textbf{0.454} & \textbf{7.987} & \textbf{1.995} & \textbf{1.058} \\ 
    \bottomrule[1.2pt]
  \end{tabular}
  \label{tab:ablation_alpha}
\end{table}

\subsubsection{Qualitative Analysis}\label{Qualitative-Analysis}

We present a qualitative comparison with SOTA methods such as SCSNet~\cite{cho2022selection} and Se-cff~\cite{nam2022stereo} in Fig.~\ref{fig:qualitative}. The selected scenes feature common challenges in stereo matching, including occlusions, sharp boundaries, and low-texture areas. Our method demonstrates superior performance, particularly in preserving boundary sharpness and reducing noise, as highlighted by the finer details and clearer transitions in the yellow boxes. In low-texture regions, our model effectively suppresses artifacts and produces coherent disparity maps, whereas competing methods tend to exhibit fragmented or noisy outputs, may due to insufficient context modeling. The magma color visualization further emphasizes our model’s ability to recover accurate depth estimations. 

\subsection{Efficiency Evaluation}
To better assess the computational characteristics of URNet, we compare its efficiency with Se-cff~\cite{nam2022stereo} and SCSNet~\cite{cho2022selection} under identical hardware conditions. We report standard metrics including floating-point operations (FLOPs), the number of trainable parameters (Params), average inference time, frames per second (FPS), and peak memory usage (Max Mem). All models are evaluated using their official implementations on an NVIDIA RTX 3090 GPU with a batch size of 1, and results are summarized in Tab.~\ref{tab:efficiency}.
URNet requires 540.49 GFLOPs and contains 7.96 million parameters, with an average inference time of 0.0700 s per frame and a peak memory usage of 719.78~MB. In comparison, Se-cff achieves faster inference (17.57 FPS) with lower FLOPs and memory consumption, but at the cost of reduced accuracy, as shown in Tab.~\ref{tab:online_results}. SCSNet demands over four times more FLOPs and nearly twice the memory compared with the URNet. These results suggest that URNet achieves a favorable trade-off between computational efficiency and predictive quality.

\subsection{Ablation Studies}\label{ablation-studies}

% We conduct ablation experiments in an offline setting, training each configuration for 100 epochs on two NVIDIA RTX 3090 GPUs.
As summarized in Tab.~\ref{tab:ablation_modules}, we assess the contribution of key components in our framework: the local-global refinement module and the uncertainty-aware loss (with $\alpha$ = 2.0). Incorporating the local refinement module results in a modest improvement, reducing the MAE to 0.472. Replacing the standard smooth $L_{1}$ loss with our uncertainty-aware loss yields a more substantial gain, lowering the MAE to 0.454 and the 1PE to 7.987\%, which highlights the benefit of explicitly modeling prediction uncertainty. When all components are combined, the model achieves the best overall performance, with 0.451 MAE, 7.896\% 1PE, and 1.057 RMSE, demonstrating the complementary advantages of refinement and uncertainty modeling.
To investigate the impact of the uncertainty weighting parameter  \(\alpha\), we keep the model architecture fixed and vary  \(\alpha\) from 1.0 to 2.0. As presented in Tab.~\ref{tab:ablation_alpha}, performance consistently improves with increasing  \(\alpha\), reaching an optimal value at  \(\alpha=2.0\). This trend suggests that appropriately weighting uncertainty enhances the model’s ability. Overall, these findings underscore the complementary advantages of our architectural design and uncertainty-aware loss in improving event-based stereo depth estimation.

\section{Conclusion}\label{sec:conclusion}

We propose the URNet, an event-based stereo depth estimation framework that integrates multi-scale disparity refinement with uncertainty modeling. This unified design effectively captures fine-grained structural details, resulting in superior prediction accuracy under challenging conditions. Extensive evaluations on the DSEC dataset show that the URNet consistently outperforms SOTA methods across multiple metrics, including MAE, 1PE, 2PE, and RMSE. Qualitative results further highlight its ability to produce sharper disparity boundaries and more stable predictions. By effectively reducing errors, the URNet demonstrates strong potential for real-world autonomous driving applications.

\paragraph{Limitations and future directions.}
While URNet achieves a favorable trade-off between accuracy and efficiency, it still incurs moderately higher computational cost than ultra-lightweight models. In future work, we plan to explore architecture simplification and model compression to reduce overhead. We also aim to extend URNet to multi-modal fusion with complementary inputs such as RGB or LiDAR, and to evaluate its generalization across diverse event-based datasets.

\vspace{.3in} 
\noindent 
\textbf{Abbreviations:} 
CNN: convolutional neural network; 
CSA: cross-scale aggregation; 
DSEC: a stereo event camera dataset for driving scenarios; 
FLOPs: floating-point operations; 
FPS: frames per second; 
FRN: feature refinement network; 
ISA: intra-scale aggregation; 
KL: kullback–leibler; 
LiDAR: light detection and ranging; 
MAE: mean absolute error; 
Max Mem: peak memory usage;  
Params: number of learnable parameters; 
PE: pixel error; 
RMSE: root mean square error; 
SCSNet: selection and cross similarity network; 
SOTA: state-of-the-art; 
URNet: uncertainty-aware refinement network.

\begin{small}
\vspace{.3in} \noindent \textbf{Data Availability:}
The datasets generated during and/or analyzed during the current study are available on the DSEC website, \url{https://dsec.ifi.uzh.ch/}. The source code of URNet is available at \href{https://github.com/yifeng-cheng/URNet}{https://github.com/yifeng-cheng/URNet}.

\vspace{.3in} \noindent \textbf{Competing Interests:}
Hu Cao is an Associate Editor at Visual Intelligence and was not involved in the editorial review of this article or the decision to publish it. The authors declare that they have no other competing interests.

\vspace{.3in} \noindent \textbf{Author Contributions:}
All authors contributed to the study conception and design. Material preparation, data collection, and analysis were performed by Yifeng Cheng, Alois Knoll, and Hu Cao. The first draft of the manuscript was written by Yifeng Cheng, and all authors commented on previous versions of the manuscript. All authors read and approved the final manuscript.

Specific contributions are as follows: 
Conceptualization: Alois Knoll, Hu Cao; 
Methodology: Hu Cao, Yifeng Cheng;  
Software: Yifeng Cheng; 
Validation: Yifeng Cheng; 
Formal analysis: Yifeng Cheng; 
Investigation: Hu Cao, Yifeng Cheng; 
Resources: Alois Knoll, Hu Cao; 
Data curation: Yifeng Cheng; 
Writing – original draft: Yifeng Cheng; 
Writing – review and editing: Hu Cao, Yifeng Cheng; 
Visualization: Yifeng Cheng. 

\end{small}

% acknowledgments part
\vspace{.3in} \noindent \textbf{Acknowledgements:}
We acknowledge the Chair of Robotics, Artificial Intelligence, and Real-time Systems at the Technical University of Munich (TUM) for providing the necessary resources and support throughout the study.

\vspace{.3in} \noindent \textbf{Funding:} 
This work is supported by the MANNHEIM-CeCaS (No. 16ME0820).

% BibTeX from reference.bib
%\bibliographystyle{sn-apacite}
%\bibliographystyle{apalike}
\bibliographystyle{unsrt}
\bibliography{reference}

\end{document}